\documentclass[10pt,twocolumn,letterpaper]{article}

\usepackage[pagenumbers]{cvpr} 

\usepackage[accsupp]{axessibility}
\usepackage{graphicx}
\usepackage{amsmath}
\usepackage{amssymb}
\usepackage{booktabs}
\usepackage{times}
\usepackage{epsfig}
\usepackage{graphicx}
\usepackage{tabularx}
\usepackage{arydshln}
\usepackage{bm}
\usepackage{nicefrac}
\usepackage{fixltx2e}
\usepackage{multirow}
\usepackage{mathtools}
\usepackage{algorithm}
\usepackage{algpseudocode}
\usepackage{breqn}
\makeatletter
\@namedef{ver@everyshi.sty}{}
\makeatother
\usepackage{tikz}
\usepackage{pifont}

\usepackage{tikz}
\usetikzlibrary{arrows,shapes,calc,matrix,fit,backgrounds}
\usepackage{pgfplots}
\usepackage{pgfplotstable}
\pgfplotsset{compat=1.9}
\usepgfplotslibrary{fillbetween}

\usepackage{xstring}

\usepgfplotslibrary{external}

\IfBeginWith*{\jobname}{fig/extern/}{\finalcopy}{}

\tikzset{every mark/.append style={solid}}
\pgfplotsset{
	grid=both, width=\columnwidth, try min ticks=5,
	every axis/.append style={font=\scriptsize},
	every axis plot/.append style={thick,mark=none,mark size=1.2,tension=0.18},
	legend cell align=left, legend style={fill opacity=0.8},
}

\pgfplotsset{
	dash/.style={mark=o,dashed,opacity=0.7},
	dott/.style={mark=o,dotted,opacity=0.7},
}

\newcommand{\xmark}{\ding{55}}
\newcommand{\ymark}{\ding{51}}

\usepackage[pagebackref,breaklinks,colorlinks]{hyperref}

\usepackage[capitalize]{cleveref}
\crefname{section}{Sec.}{Secs.}
\Crefname{section}{Section}{Sections}
\Crefname{table}{Table}{Tables}
\crefname{table}{Tab.}{Tabs.}

\def\cvprPaperID{5968} 
\def\confName{CVPR}
\def\confYear{2023}

\renewcommand{\paragraph}[1]{\noindent\textbf{#1}}

\begin{document}

\title{SimCon Loss with Multiple Views for Text Supervised Semantic Segmentation}

\author{Yash Patel$^{1}$ \thanks{The research was conducted during Y. Patel’s internship at AWS.}, \quad Yusheng Xie$^{2}$ \thanks{Corresponding author.}, \quad Yi Zhu$^{2}$, \quad Srikar Appalaraju$^{2}$, \quad R. Manmatha$^{2}$ \\
$^{1}$ Visual Recognition Group, Czech Technical University in Prague \quad
$^{2}$ AWS AI Labs\\
{\tt\small patelyas@fel.cvut.cz, \{yushx, yzaws, srikara, manmatha\}@amazon.com}
}

\maketitle
\newcommand{\nn}[1]{\ensuremath{\text{NN}_{#1}}\xspace}
\def\l1{\ensuremath{\ell_1}\xspace}
\def\l2{\ensuremath{\ell_2}\xspace}

\def\roxf{$\mathcal{R}$Oxford\xspace}
\def\rox{$\mathcal{R}$Oxf\xspace}
\def\ro{$\mathcal{R}$O\xspace}
\def\rpar{$\mathcal{R}$Paris\xspace}
\def\rpa{$\mathcal{R}$Par\xspace}
\def\rp{$\mathcal{R}$P\xspace}
\def\rdis{$\mathcal{R}$1M\xspace}

\newcommand\resnet[3]{\ensuremath{\prescript{#2}{}{\mathtt{R}}{#1}_{\scriptscriptstyle #3}}\xspace}

\newcommand*\OK{\ding{51}}

\newenvironment{narrow}[1][1pt]
	{\setlength{\tabcolsep}{#1}}
	{\setlength{\tabcolsep}{6pt}}

\newcommand{\alert}[1]{{\color{red}{#1}}}
\newcommand{\gio}[1]{{\color{blue}{#1}}}
\newcommand{\replace}[2]{{\color{gray}{#1}}{\color{red}{#2}}}

\newcommand{\comment} [1]{{\color{orange} \Comment     #1}} 


\newcommand{\head}[1]{{\smallskip\noindent\bf #1}}
\newcommand{\equ}[1]{(\ref{equ:#1})\xspace}

\newcommand{\red}[1]{{\color{red}{#1}}}
\newcommand{\blue}[1]{{\color{blue}{#1}}}
\newcommand{\green}[1]{{\color{green}{#1}}}
\newcommand{\gray}[1]{{\color{gray}{#1}}}


\newcommand{\tran}{^\top}
\newcommand{\mtran}{^{-\top}}
\newcommand{\zcol}{\mathbf{0}}
\newcommand{\zrow}{\zcol\tran}

\newcommand{\ind}{\mathds{1}}
\newcommand{\expect}{\mathbb{E}}
\newcommand{\nat}{\mathbb{N}}
\newcommand{\zahl}{\mathbb{Z}}
\newcommand{\real}{\mathbb{R}}
\newcommand{\proj}{\mathbb{P}}
\newcommand{\prob}{\mathbf{Pr}}

\newcommand{\mif}{\textrm{if }}
\newcommand{\other}{\textrm{otherwise}}
\newcommand{\minimize}{\textrm{minimize }}
\newcommand{\maximize}{\textrm{maximize }}

\newcommand{\id}{\operatorname{id}}
\newcommand{\const}{\operatorname{const}}
\newcommand{\sgn}{\operatorname{sgn}}
\newcommand{\var}{\operatorname{Var}}
\newcommand{\mean}{\operatorname{mean}}
\newcommand{\trace}{\operatorname{tr}}
\newcommand{\diag}{\operatorname{diag}}
\newcommand{\vect}{\operatorname{vec}}
\newcommand{\cov}{\operatorname{cov}}

\newcommand{\softmax}{\operatorname{softmax}}
\newcommand{\clip}{\operatorname{clip}}

\newcommand{\defn}{\mathrel{:=}}
\newcommand{\peq}{\mathrel{+\!=}}
\newcommand{\meq}{\mathrel{-\!=}}

\newcommand{\floor}[1]{\left\lfloor{#1}\right\rfloor}
\newcommand{\ceil}[1]{\left\lceil{#1}\right\rceil}
\newcommand{\inner}[1]{\left\langle{#1}\right\rangle}
\newcommand{\norm}[1]{\left\|{#1}\right\|}
\newcommand{\frob}[1]{\norm{#1}_F}
\newcommand{\card}[1]{\left|{#1}\right|\xspace}
\newcommand{\diff}{\mathrm{d}}
\newcommand{\der}[3][]{\frac{d^{#1}#2}{d#3^{#1}}}
\newcommand{\pder}[3][]{\frac{\partial^{#1}{#2}}{\partial{#3^{#1}}}}
\newcommand{\ipder}[3][]{\partial^{#1}{#2}/\partial{#3^{#1}}}
\newcommand{\dder}[3]{\frac{\partial^2{#1}}{\partial{#2}\partial{#3}}}

\newcommand{\wb}[1]{\overline{#1}}
\newcommand{\wt}[1]{\widetilde{#1}}

\def\nsp{\hspace{-3pt}}
\def\zsp{\hspace{0pt}}
\def\xssp{\hspace{1pt}}
\def\ssp{\hspace{3pt}}
\def\msp{\hspace{6pt}}
\def\lsp{\hspace{12pt}}
\def\xlsp{\hspace{20pt}}

\newcommand{\cA}{\mathcal{A}}
\newcommand{\cB}{\mathcal{B}}
\newcommand{\cC}{\mathcal{C}}
\newcommand{\cD}{\mathcal{D}}
\newcommand{\cE}{\mathcal{E}}
\newcommand{\cF}{\mathcal{F}}
\newcommand{\cG}{\mathcal{G}}
\newcommand{\cH}{\mathcal{H}}
\newcommand{\cI}{\mathcal{I}}
\newcommand{\cJ}{\mathcal{J}}
\newcommand{\cK}{\mathcal{K}}
\newcommand{\cL}{\mathcal{L}}
\newcommand{\cM}{\mathcal{M}}
\newcommand{\cN}{\mathcal{N}}
\newcommand{\cO}{\mathcal{O}}
\newcommand{\cP}{\mathcal{P}}
\newcommand{\cQ}{\mathcal{Q}}
\newcommand{\cR}{\mathcal{R}}
\newcommand{\cS}{\mathcal{S}}
\newcommand{\cT}{\mathcal{T}}
\newcommand{\cU}{\mathcal{U}}
\newcommand{\cV}{\mathcal{V}}
\newcommand{\cW}{\mathcal{W}}
\newcommand{\cX}{\mathcal{X}}
\newcommand{\cY}{\mathcal{Y}}
\newcommand{\cZ}{\mathcal{Z}}

\newcommand{\vA}{\mathbf{A}}
\newcommand{\vB}{\mathbf{B}}
\newcommand{\vC}{\mathbf{C}}
\newcommand{\vD}{\mathbf{D}}
\newcommand{\vE}{\mathbf{E}}
\newcommand{\vF}{\mathbf{F}}
\newcommand{\vG}{\mathbf{G}}
\newcommand{\vH}{\mathbf{H}}
\newcommand{\vI}{\mathbf{I}}
\newcommand{\vJ}{\mathbf{J}}
\newcommand{\vK}{\mathbf{K}}
\newcommand{\vL}{\mathbf{L}}
\newcommand{\vM}{\mathbf{M}}
\newcommand{\vN}{\mathbf{N}}
\newcommand{\vO}{\mathbf{O}}
\newcommand{\vP}{\mathbf{P}}
\newcommand{\vQ}{\mathbf{Q}}
\newcommand{\vR}{\mathbf{R}}
\newcommand{\vS}{\mathbf{S}}
\newcommand{\vT}{\mathbf{T}}
\newcommand{\vU}{\mathbf{U}}
\newcommand{\vV}{\mathbf{V}}
\newcommand{\vW}{\mathbf{W}}
\newcommand{\vX}{\mathbf{X}}
\newcommand{\vY}{\mathbf{Y}}
\newcommand{\vZ}{\mathbf{Z}}

\newcommand{\va}{\mathbf{a}}
\newcommand{\vb}{\mathbf{b}}
\newcommand{\vc}{\mathbf{c}}
\newcommand{\vd}{\mathbf{d}}
\newcommand{\ve}{\mathbf{e}}
\newcommand{\vf}{\mathbf{f}}
\newcommand{\vg}{\mathbf{g}}
\newcommand{\vh}{\mathbf{h}}
\newcommand{\vi}{\mathbf{i}}
\newcommand{\vj}{\mathbf{j}}
\newcommand{\vk}{\mathbf{k}}
\newcommand{\vl}{\mathbf{l}}
\newcommand{\vm}{\mathbf{m}}
\newcommand{\vn}{\mathbf{n}}
\newcommand{\vo}{\mathbf{o}}
\newcommand{\vp}{\mathbf{p}}
\newcommand{\vq}{\mathbf{q}}
\newcommand{\vr}{\mathbf{r}}
\newcommand{\Vs}{\mathbf{s}}
\newcommand{\vt}{\mathbf{t}}
\newcommand{\vu}{\mathbf{u}}
\newcommand{\vv}{\mathbf{v}}
\newcommand{\vw}{\mathbf{w}}
\newcommand{\vx}{\mathbf{x}}
\newcommand{\vy}{\mathbf{y}}
\newcommand{\vz}{\mathbf{z}}

\newcommand{\vone}{\mathbf{1}}
\newcommand{\vzero}{\mathbf{0}}

\newcommand{\valpha}{{\boldsymbol{\alpha}}}
\newcommand{\vbeta}{{\boldsymbol{\beta}}}
\newcommand{\vgamma}{{\boldsymbol{\gamma}}}
\newcommand{\vdelta}{{\boldsymbol{\delta}}}
\newcommand{\vepsilon}{{\boldsymbol{\epsilon}}}
\newcommand{\vzeta}{{\boldsymbol{\zeta}}}
\newcommand{\veta}{{\boldsymbol{\eta}}}
\newcommand{\vtheta}{{\boldsymbol{\theta}}}
\newcommand{\viota}{{\boldsymbol{\iota}}}
\newcommand{\vkappa}{{\boldsymbol{\kappa}}}
\newcommand{\vlambda}{{\boldsymbol{\lambda}}}
\newcommand{\vmu}{{\boldsymbol{\mu}}}
\newcommand{\vnu}{{\boldsymbol{\nu}}}
\newcommand{\vxi}{{\boldsymbol{\xi}}}
\newcommand{\vomikron}{{\boldsymbol{\omikron}}}
\newcommand{\vpi}{{\boldsymbol{\pi}}}
\newcommand{\vrho}{{\boldsymbol{\rho}}}
\newcommand{\vsigma}{{\boldsymbol{\sigma}}}
\newcommand{\vtau}{{\boldsymbol{\tau}}}
\newcommand{\vupsilon}{{\boldsymbol{\upsilon}}}
\newcommand{\vphi}{{\boldsymbol{\phi}}}
\newcommand{\vchi}{{\boldsymbol{\chi}}}
\newcommand{\vpsi}{{\boldsymbol{\psi}}}
\newcommand{\vomega}{{\boldsymbol{\omega}}}

\newcommand{\rLambda}{\mathrm{\Lambda}}
\newcommand{\rSigma}{\mathrm{\Sigma}}

\makeatletter
\DeclareRobustCommand\onedot{\futurelet\@let@token\@onedot}
\def\@onedot{\ifx\@let@token.\else.\null\fi\xspace}
\def\eg{\emph{e.g}\onedot} \def\Eg{\emph{E.g}\onedot}
\def\ie{\emph{i.e}\onedot} \def\Ie{\emph{I.e}\onedot}
\def\cf{\emph{cf}\onedot} \def\Cf{\emph{C.f}\onedot}
\def\etc{\emph{etc}\onedot} \def\vs{\emph{vs}\onedot}
\def\wrt{w.r.t\onedot} \def\dof{d.o.f\onedot}
\def\etal{\emph{et al}\onedot}
\makeatother

\begin{abstract}
Learning to segment images purely by relying on the image-text alignment from web data can lead to sub-optimal performance due to noise in the data. The noise comes from the samples where the associated text does not correlate with the image's visual content. Instead of purely relying on the alignment from the noisy data, this paper proposes a novel loss function termed SimCon, which accounts for intra-modal similarities to determine the appropriate set of positive samples to align. Further, using multiple views of the image (created synthetically) for training and combining the SimCon loss with it makes the training more robust. This version of the loss is termed MV-SimCon. The empirical results demonstrate that using the proposed loss function leads to consistent improvements on zero-shot, text supervised semantic segmentation and outperforms state-of-the-art by $+3.0\%$, $+3.3\%$ and $+6.9\%$ on PASCAL VOC, PASCAL Context and MSCOCO, respectively. With test time augmentations, we set a new record by improving these results further to $58.7\%$, $26.6\%$, and $33.3\%$ on PASCAL VOC, PASCAL Context, and MSCOCO, respectively. In addition, using the proposed loss function leads to robust training and faster convergence. \footnote{Code and the trained models will be publicly released.}

\end{abstract}

\section{Introduction}
\label{sec:introduction}

\begin{figure}[t]
\begin{center}
\includegraphics[width=0.42\textwidth]{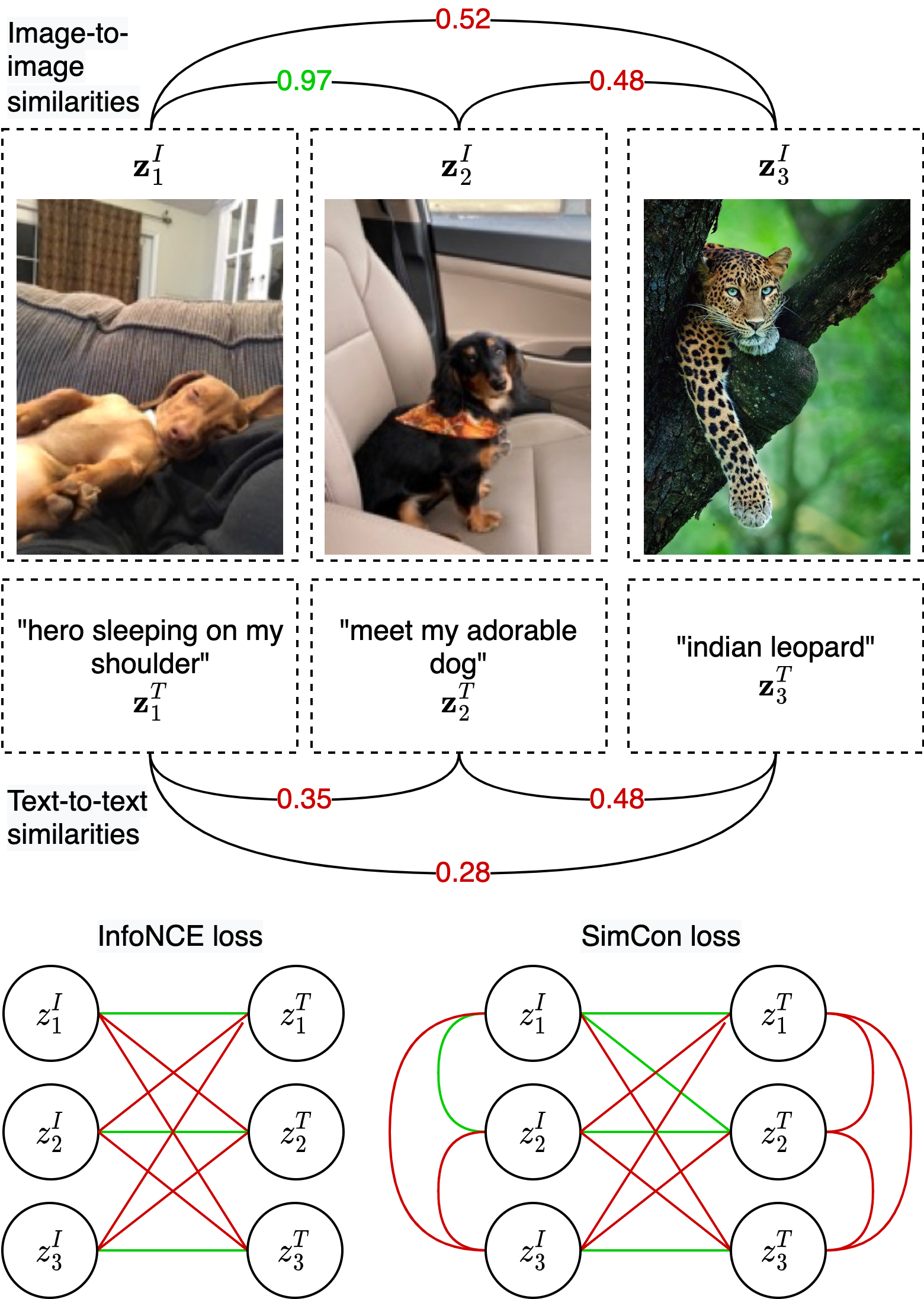}
\caption{\textbf{Top}: In the above image-text samples, the text corresponding to the first image does not contain any information about its visual content. The text for the second image correctly describes the visual content of both the first and the second images. {\textbf{Bottom}: If we use the InfoNCE loss, only the image and its paired text will be pulled together (\textcolor{green}{green} edges), while all other images and captions will be pushed apart (\textcolor{red}{red} edges). In SimCon loss, both the first and images will be paired with the text from the second image and pushed away from the first piece of text. The loss function learns this automatically. Best viewed in color.}}
\label{fig:teaser}
\end{center}
\vspace{-3em}
\end{figure}

The use of data from the web for training visual and language models has been effective for learning visual representations that are useful for downstream tasks. Learning the visual and textual encoders jointly via projecting images and texts to the same learned embedding space allows their direct comparison. It helps in developing open-set and zero-shot classification models \cite{clip-icml21,align-icml21,alayrac2022flamingo,florence-arxiv21,pali-arxiv22,coca-arxiv22,singh2022flava,yao2021filip,zhai2022lit,pham2021combined}. This success has prompted researchers to investigate the use of web data for learning object-level representations for tasks involving dense predictions such as semantic segmentation \cite{xu2022groupvit}, without fine-tuning on any dense supervision. Whether the focus is on learning global or object-level representations, these approaches rely on the alignment between the images and the co-occurring text for supervision.

The cross-modal alignment for zero-shot image classification or segmentation models is performed via contrastive learning using InfoNCE \cite{hcl06,cpc} loss function, which maximizes the mutual information between the image and its matched text. The loss is computed by using each batch sample as an anchor. In the embedding space, for an anchor image (text), the embedding vector of the corresponding text (image) is pulled closer. In contrast, the embedding vectors of texts (images) from other samples are pushed apart. While this objective has been useful \cite{clip-icml21,align-icml21,xu2022groupvit}, it is prone to noise in the training data, which is typical for samples from the web. Often the text associated with an image does not describe its visual content, may miss the information for objects in the background, or could be ambiguous. As shown in \cref{fig:teaser}, the caption associated with the first image does not describe the visual content, whereas the caption for the second image describes the visual content of both the first and the second image. Using the InfoNCE loss function for these samples will push apart the embedding vectors of the first image and the second text, leading to sub-optimal representations.

This work focuses on learning a model for text supervised zero-shot semantic segmentation without using any weakly supervised or dense annotations. This paper builds upon the only existing baseline for the task, {\em i.e.}, GroupViT \cite{xu2022groupvit}, which suffers from the aforementioned noisy training. To mitigate the issue, this paper proposes a new intra-modal similarity aware contrastive loss function termed \text{\em SimCon}. As shown in \cref{fig:teaser}, the computation of SimCon starts by computing the intra-modal image-to-image and text-to-text similarities. For an image as an anchor, a set of positive image samples are assigned if the intra-modal image-to-image similarity is higher than a threshold. For the anchor image, the SimCon loss then pulls closer the embedding vectors of the corresponding text, positive image samples, and their corresponding texts and pushes apart the embedding vectors of the remaining images and texts. Similarly, the positives for a text as an anchor are determined based on the intra-modal text-to-text similarities, and the SimCon loss pulls the embedding vectors of the corresponding image, positive text samples, and their corresponding images and pushes apart the rest of the text and image embedding vectors.

For visual representation learning, several approaches also use InfoNCE to pull together the embedding vectors from different synthetically generated views of the same image and push apart the embedding vectors from different images. They have shown to be useful for both self-supervised \cite{chen2021exploring,he2020momentum,chen2021empirical,caron2020unsupervised,caron2021emerging} and supervised learning \cite{khosla2020supervised}. While SimCon loss, as shown in \cref{fig:teaser} already accounts for intra-modal relations using the intra-modal similarities, it is methodically extended to account for multiple views of the image, and the setup is termed \text{\em MV-SimCon} for brevity. With these systematic improvements, this paper sets a new state-of-the-art for zero-shot semantic segmentation that trains without manually annotated data.  

The contributions of this paper are as follows:
\begin{itemize}
 \item A novel {\em SimCon} loss with multiple views is introduced to mitigate the issue of noisy image-text pair training.
 \item The proposed loss is robust across different data distributions during training and scales with the amount of training data and batch size.
 \item Extensive empirical results demonstrate the superiority of {\em MV-SimCon} in terms of faster convergence and better zero-shot segmentation performance.
\end{itemize}
\section{Related Work}
\label{sec:related_work}

\paragraph{Semantic segmentation without dense supervision.} Semantic segmentation is a dense prediction task that assigns a semantic label to each pixel. Most methods rely on annotated data to achieve decent segmentation results \cite{chen2017deeplab,chen2017rethinking,zhao2017pyramid,long2015fully,li2022mask,zheng2021rethinking,strudel2021segmenter,cheng2021per,xie2021segformer,zhu-pami-selfseg}. Given costly pixel-wise annotations, there have been several attempts to do unsupervised semantic segmentation.  One way is to learn representations for each pixel and then perform clustering to obtain dense semantic labels. Earlier work~\cite{ji-iccv2019-iic,ouali-eccv2020-arseg} demonstrates the possibility of clustering on small-scale datasets.  To facilitate object discovery, recent work has started to incorporate more prior information~\cite{cho-cvpr2021-PiCIE,gansbeke-iccv2021-maskcontrast,gansbeke-arxiv2022-maskdistill} or better image representations~\cite{hamilton-iclr2022-stego,zadaianchuk-arxiv2022-comus,ziegler-cvpr2022-leopart}. Yet the performance still lags its supervised counterparts. Another way is to use language supervision as a weak signal. Several recent papers~\cite{simplebaseline, lseg,rao2021denseclip,clipseg,zhou2022extract} leverage the CLIP model~\cite{clip-icml21} to enable open-set semantic segmentation, but they also require dense supervision. OpenSeg~\cite{openseg}  goes one step further by using image-level supervision to learn with class-agnostic mask annotation. Finally, GroupViT~\cite{xu2022groupvit} shows that semantic segmentation can be done by training a model on web data image-text pairs without mask annotations. In this work, we instantiate our framework using GroupViT but change the training objectives from InfoNCE to the proposed MV-SimCon. The goal is to mitigate the noisy image-text supervision in GroupViT training. 

\paragraph{Learning visual representations from web data.} Web data as a source of supervision has been a promising direction to learn visual representations~\cite{webvision,patel2018texttopicnet,patel2019self,gomez2017self,gordo2017beyond}. With the help of metadata like tags and alt-text, the labeling cost of such datasets can be reduced significantly, which leads to cheaper large-scale datasets. In order to study the effect of data in the deep learning era, YFCC100M~\cite{yfcc100m}, JFT300M~\cite{jft300m-iccv17}, JFT3B~\cite{jft3b-cvpr22}, IG3.5B~\cite{ig35b-eccv18} and  IG65M~\cite{ig65m-cvpr19} were collected and studied. As expected, larger datasets help to learn better visual representations and lead to state-of-the-art results for various vision tasks. With the rise of multi-modality learning, there is a recent trend of using image-text pairs from the web as supervision~\cite{clip-icml21,align-icml21,florence-arxiv21,coca-arxiv22,beitv3-arxiv22}. Thanks to the larger datasets~\cite{laion400m} and larger transformer models~\cite{jft3b-cvpr22,pali-arxiv22}, these trained models have capabilities, such as zero-shot prediction. 

\paragraph{Contrastive loss objectives.} Contrastive losses have been applied to a wide variety of data from several domains, e.g., computer vision~\cite{chen2021exploring,he2020momentum,chen2021empirical,caron2020unsupervised}, natural language processing~\cite{mrasp2,gao2021simcse}, speech and audio~\cite{cpc,cola} and multi-modal~\cite{clip-icml21,align-icml21}. These losses can be used as long as the anchor, positives, and negatives are well-defined. Such losses have widely been studied in the open-set image retrieval literature \cite{mbl20}, where the deep embedding is trained on a set of classes, and the retrieval evaluation is performed on unseen classes from the same distribution. The simplest pairwise loss function is the contrastive loss \cite{hcl06}, also known as InfoNCE, where the embeddings of the relevant pair of samples are pulled as close as possible, and the non-relevant ones are pushed apart. The triplet loss \cite{spk+14,weinberger2009distance} mimics a ranking objective more closely by training on a triplet pair of an anchor, a positive and a negative sample. Since optimization over all possible combination of samples is not tractable, much attention has been paid to finding informative pairs via sampling \cite{sxj+15,sohn16,rms+20,wms+17,lxz+19,bxk+20,rmp+20,chx+19,patel2022recall,zolfaghari2021crossclr}. All the above-mentioned loss functions involving approximation of the evaluation metric or sampling have been studied in a uni-modal, supervised setup with class-balance sampling, which is not feasible with the multi-modal data from the web without any semantic labels. Therefore, the standard InfoNCE \cite{hcl06} has been widely adopted to date due to its simplicity. Our work takes motivation from the image retrieval approaches and attempts to find a better sampling strategy for multi-modal contrastive learning, in the absence of semantic labels. We incorporate intra-modal similarities and multiple views to determine adequate positive samples in the noisy data from the web.

\paragraph{Vision-language training with noisy data.} Recent approaches, designed mainly for cross-modal retrieval such as ALBEF \cite{li2021align}, TCL \cite{yang2022vision}, CodeBook \cite{duan2022multi}, and BLIP \cite{li2022blip} also attempt to mitigate noise during training in various ways. While \cite{clip-icml21,align-icml21,xu2022groupvit} use separate text and image encoders, \cite{li2021align,yang2022vision,duan2022multi,li2022blip} also use a multi-modal encoder. For noise mitigation, \cite{li2021align,yang2022vision,duan2022multi} use distillation through an exponentially moving average momentum encoder, whereas \cite{li2022blip} curates the training data using additional captioning and filtering models. These approaches are computationally expensive as they require additional models, have only been investigated for tasks requiring global predictions, such as cross-modal retrieval, where their efficacy is marginal.

\begin{figure*}[t]
\begin{center}
\includegraphics[width=\linewidth]{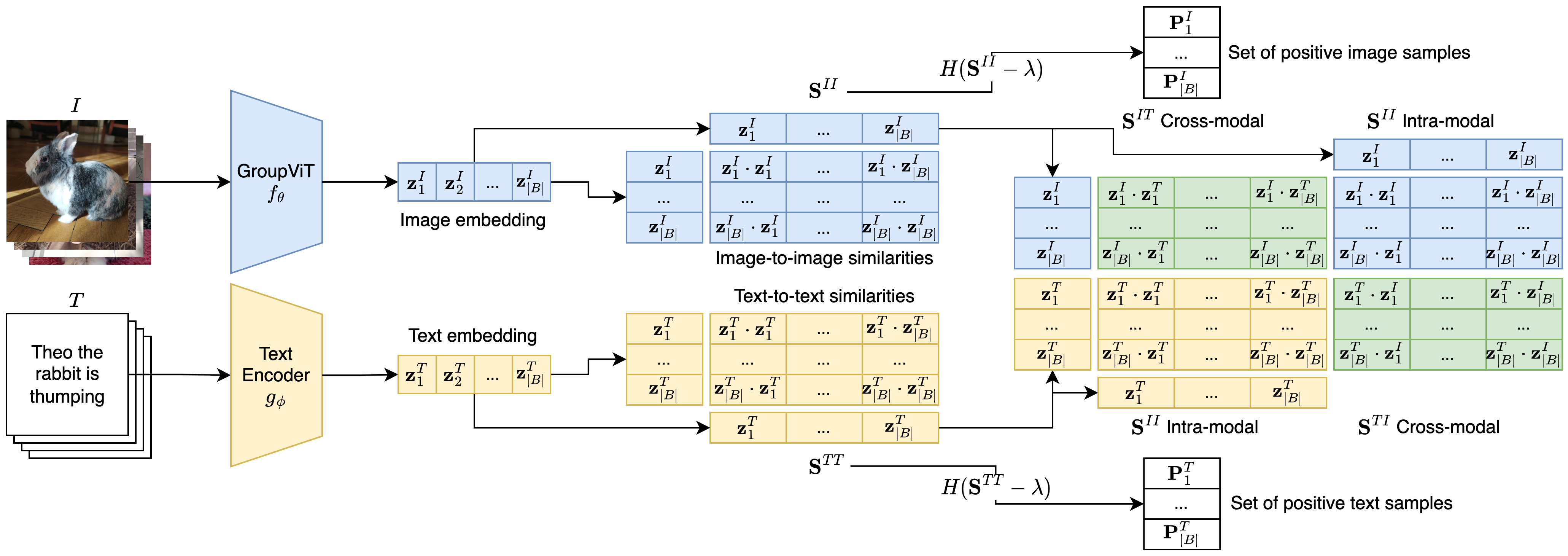}
\vspace{-2em}
\caption{\textbf{SimCon Overview.} During training, the sampled images $I$ are passed through the GroupViT model $f_{\theta}$, and the segment tokens are averaged and normalized to obtain the embedding $\vz^{I}$. The texts $T$ are passed through the text encoder $g_{\phi}$ to obtain the text embedding $\vz^{T}$. The intra-modal image-to-image $\vS^{II}$ and text-to-text $\vS^{TT}$ similarities are computed via the cosine distance. The set of positives $\vP^{I}$ and $\vP^{T}$ are determined using the intra-modal similarities by finding the samples with a similarity higher than the threshold $\lambda$. This is achieved by passing $\vS^{II} - \lambda$ and $\vS^{TT} - \lambda$ through a Heaviside step function $H$, the output $1$ determines positives and $0$ for negatives. The SimCon loss defined in \cref{eq:simcon_im2txt} and (\ref{eq:simcon_txt2im}) is computed on a joint similarity matrix containing both the intra-modal and cross-modal similarities with the positive and negative relations between the pairs  governed by $\vP^{I}$ and $\vP^{T}$. \textcolor{blue}{Blue} shows image modality, \textcolor{yellow}{yellow} shows text modality and \textcolor{green}{green} shows cross-modal. MV-SimCon follows a similar pipeline with multiple views as elaborated in \cref{subsec:mvsimcon}.
}
\label{fig:overview}
\vspace{-2em}
\end{center}
\end{figure*}

\section{Method}
\label{sec:method}
This section revisits GroupViT as a baseline to describe its model architecture and training objectives (\cref{subsec:groupvit}). Then we introduce our improved contrastive losses by bringing intra-modal similarity (SimCon) (\cref{subsec:simcon}) and multi-view (MV-SimCon) into the picture (\cref{subsec:mvsimcon}).

\subsection{Preliminary}
\label{subsec:groupvit}

\paragraph{GroupViT.} GroupViT \cite{xu2022groupvit} is the first work to explore zero-shot transfer from text supervision alone to  semantic segmentation  without using any pixel-wise labels. The basic idea is to bring back the grouping mechanism \cite{Yu2002ConcurrentOR,Shi1997NormalizedCA,Yu2001GroupingWB} into deep transformer networks in a bottom-up manner. Through a hierarchical grouping process, the model learns to grow image regions into progressively larger arbitrary-shaped segments.

To be specific, GroupViT consists of a vision encoder $f_{\theta}$ and a text encoder $g_{\phi}$. The vision encoder is a vision transformer (ViT) with group tokens and grouping block. Given a batch of images and the corresponding texts $\{(I_{i}, T_{i})\}$, where $i$ is the index in the batch size, the batch is sampled from a collection of multi-modal data. During feed-forward, the image is first split into non-overlapping patches and linearly projected into a latent space, which are termed as segment tokens. The segment tokens and the learnable grouping tokens are then fed to the transformer layers. After a set of transformer layers, the segment tokens and the grouping tokens are passed to a grouping block. Within the grouping block, segment tokens are assigned to groups and merged together for further processing. The assignment is done by computing the similarities between the segment tokens and the group tokens and using a differentiable Gumbel-softmax assignment \cite{jang2016categorical,maddison2016concrete}. The merging combines all the segment tokens belonging to the same group and is performed via a weighted sum. In this way, the group tokens can thus learn to aggregate information globally from all segment tokens. The set of transformer layers followed by a grouping block constitutes a stage. Stacking two such stages stacked together gives the final vision encoder. For training, a global representation of the image $\{\vz^{I}_{i}\} \in \real^d$ is obtained by average pooling the final segment tokens, followed by $L_{2}$-norm. The text encoder is a transformer model and is the same as in~\cite{clip-icml21} and the final normalized text embeddings are denoted as $\{\vz^{T}_{i}\} \in \real^d$.

\paragraph{Notations.}
Usually the similarity between any two embedding vectors is computed by the dot product between them and is denoted by $s(\vz^{I}_{i}, \vz^{T}_{i})=\vz^{I}_{i} \cdot \vz^{T}_{i}$. Within the batch $B=\{(\vz^{I}_{i}, \vz^{T}_{i})\}$, the similarity between all images and texts is computed and is stored in a $|B| \times |B|$ dimensional matrix $\vS^{IT}$. For brevity, $\vS^{IT}_{ij}$ is the similarity between an image with index $i$ and text with index $j$. Similarly a matrix $\vS^{II}$ contains the image-to-image similarities and $\vS^{TT}$ text-to-text similarities. The computation of InfoNCE loss  involves the temperature controlled exponent of the similarities,and it is represented as $\vE^{IT}_{ij} = \exp (\vz_{i}^{I} \cdot \vz_{j}^{T}/\tau) = \exp(\vS^{IT}_{ij}/\tau)$. Here, $\tau$ is a learnable temperature parameter initialized with a value of $0.07$ \cite{clip-icml21,xu2022groupvit}. Similarly the exponent term between two images is represented as $\vE^{II}_{ij}$ and between two pieces of text as $\vE^{TT}_{ij}$.

\paragraph{InfoNCE loss.}
With a global representation for both image and text modalities, GroupViT uses InfoNCE loss function that matches an image to the corresponding text and vice versa. InfoNCE loss pulls the representations of the corresponding image and text pairs closer and pushes the representations of non-matching (according to data) text samples apart. 
This image-text alignment loss jointly trains the visual and the textual encoders, and may be expressed for image-to-text matching as: 

\begin{dmath}
    \mathcal{L}_{\text{NCE}}(I,T) = - \frac{1}{|B|} \sum_{i=1}^{|B|} \log \frac{\vE^{IT}_{ii}}{\sum_{j=1}^{|B|}\vE^{IT}_{ij}}
\label{eq:infonce_im2txt}
\end{dmath}
and similarly for text-to-image matching as:
\begin{dmath}
    \mathcal{L}_{\text{NCE}}(T,I) = - \frac{1}{|B|} \sum_{i=1}^{|B|} \log \frac{\vE^{TI}_{ii}}{\sum_{j=1}^{|B|}\vE^{TI}_{ij}}
\label{eq:infonce_txt2im}
\end{dmath}
where $\vE^{TI}$ is the transpose of $\vE^{IT}$. The overall training loss of GroupViT is a linear combination of \cref{eq:infonce_im2txt} and (\ref{eq:infonce_txt2im}).

\paragraph{Shortcomings of InfoNCE.}  Given an image and its corresponding aligned text (according to the groundtruth), InfoNCE loss pulls the image embedding closer to its corresponding text embedding and pushes the other text embeddings away. As shown in Fig.~\ref{fig:teaser}, this objective falls short in accounting for noise in the training data where the corresponding text does not contain any or partial information about the visual content of the image. Additionally, InfoNCE does not account for any intra-modal relations. The proposed SimCon loss mitigates this issue as an image is not only positively matched to the corresponding text, but also to additional images and texts determined via the intra-modal similarities.

Furthermore, the InfoNCE loss function does not account for any relations across different views of the image which has shown to be useful for  visual representation learning \cite{chen2021exploring,he2020momentum,caron2020unsupervised,chen2021empirical,caron2021emerging} and is available without any supervision, as these views are synthetically generated by applying augmentations. To add a further training signal, the SimCon loss is systematically extended to use multiple views of the images, termed MV-SimCon. Where the images in each view are positively matched to the text following the SimCon objective and an additional cosine distance loss \cite{chen2021exploring} is used to match the images from different views.

\subsection{SimCon Loss}
\label{subsec:simcon}

The proposed SimCon loss jointly accounts for intra-modal and cross-modal relations. For an image anchor, the representations of other similar images, texts of these similar images and the corresponding text to the anchor are pulled closer while the rest of the image and text representations are pushed apart. An overview of the process is shown in \cref{fig:overview}. The SimCon loss for image-to-text alignment may be expressed as:

\begin{dmath}
     \mathcal{L}_{\text{SimCon}}(I,T,\vP^{I}) \\ =  - \frac{1}{|B|} \sum_{i=1}^{|B|} \frac{1}{|\vP^{I}_{i}|}
    \sum_{p \in \vP^{I}_{i} } \log \frac{\vE^{IT}_{ip} + \vE^{II}_{ip}}{\sum_{j=1}^{|B|}\vE^{IT}_{ij} +  \sum_{j=1}^{|B|}\vE^{II}_{ij}}
\label{eq:simcon_im2txt}
\end{dmath}
and for text-to-image alignment as:
\begin{dmath}
    \mathcal{L}_{\text{SimCon}}(T,I,\vP^T) = - \frac{1}{|B|} \sum_{i=1}^{|B|} \frac{1}{|\vP^{T}_{i}|}
    \sum_{p \in \vP^{T}_{i} } \log \frac{\vE^{TI}_{ip} + \vE^{TT}_{ip}}{\sum_{j=1}^{|B|} \vE^{TI}_{ij} +  \sum_{j=1}^{|B|} \vE^{TT}_{ij}} 
\label{eq:simcon_txt2im}
\end{dmath}
Here $\vP^{I}_{i}$ is the set of images that are similar to the image anchor at $i$ and $\vP^{T}_{i}$ is the set of texts that are similar to the text anchor at $i$. As shown in \cref{fig:overview}, $\vP^{I}$ and $\vP^{T}$ are obtained from the intra-modal similarities as:

\begin{equation}
\begin{aligned}
    \vP^{I} = H(\vS^{II} - \lambda) \\
    \vP^{T} = H(\vS^{TT} - \lambda)
    \label{eqn:independent_p}
\end{aligned}
\end{equation}
where $H$ is a Heaviside step function with $H(x) = 1$ if $x>=0$, otherwise $H(x)=0$, thus the values in $\vP^{I}$ and $\vP^{T}$ are binary. $\lambda$ is a threshold hyper-parameter on the intra-modal similarities. For an image anchor $\vz^{I}_{i}$, the positive samples are the ones which have the intra-modal similarity higher than the threshold, \ie, where $\vP^{I}_{i}=1$.

\subsection{MV-SimCon: SimCon with Multiple Views}
\label{subsec:mvsimcon}

The proposed MV-SimCon loss, or SimCon loss with multiple views, enforces consistency across multiple image views obtained via data augmentation. Let $I_{1}$ and $I_{2}$ be the two views of an image, then the MV-SimCon loss for image to text alignment may be expressed as:
\begin{dmath}
    \mathcal{L}_{\text{MV-SimCon}}(I,T,\vP^{I}_{J})  =  \mathcal{L}_{\text{SimCon}}(I_1,T,\vP^{I}_{J}) + \mathcal{L}_{\text{SimCon}}(I_2,T,\vP^{I}_{J})
\label{eq:mvsimcon_im2txt}
\end{dmath}
and for text to image alignment as:
\begin{dmath}
    \mathcal{L}_{\text{MV-SimCon}}(T,I,\vP^T) = 
    \mathcal{L}_{\text{SimCon}}(T,I_1,\vP^T) + \mathcal{L}_{\text{SimCon}}(T,I_2,\vP^T)
\label{eq:mvsimcon_txt2im}
\end{dmath}
where $\vP^{I}_{J}$ is the set of images that are similar to the anchor image in either view and $\vP^T$ remains the same as in SimCon loss as no text augmentations are used. $\vP^{I}_{J}$ for the MV-SimCon loss may be expressed as:
\begin{dmath}
    \vP^{I}_{J} = H(\max(\vS^{I_1I_1}, \vS^{I_2I_2}) - \lambda)
    \label{eqn:joint_p}
\end{dmath}
where $\vS^{I_1I_1}$ are the intra-modal similarities within the first view of the images and $\vS^{I_2I_2}$ within the second view. Note that there are two possibilities, the first is to independently compute the image positives in each view and the second is to compute them jointly across the views. The joint computation of the image positives leads to more number of positives for each image and empirically leads to better performance as shown in the ablations \cref{sec:ablation}.

So far, the MV-SimCon loss aligns the images in both views to the appropriate texts following the SimCon loss. However, the images in the two views are still not connected. To connect them to each other and for an additional training signal, a negative cosine similarity loss \cite{chen2021exploring} is used between the two views of the image:
\begin{dmath}
    \mathcal{L}_{\text{NCS}}(I_{1},I_{2}) = -\frac{1}{|B|} \sum_{i=1}^{|B|} \frac{1}{2} p(\vz^{I_1}_{i}) \cdot \text{sg}(\vz^{I_2}_{i}) + \frac{1}{2} p(\vz^{I_2}_{i}) \cdot \text{sg}(\vz^{I_1}_{i})
\label{eq:ncs_loss}
\end{dmath}
where {\em sg} is the stop-gradient operation and $p$ is a projection head \cite{chen2021exploring}. The overall objective with the MV-SimCon loss is governed by a linear combination of \cref{eq:mvsimcon_im2txt}, (\ref{eq:mvsimcon_txt2im}) and (\ref{eq:ncs_loss}), 

\begin{dmath}
    \mathcal{L}_{\text{final}} = \mathcal{L}_{\text{MV-SimCon}}(I,T,\vP^{I}_{J}) + \mathcal{L}_{\text{MV-SimCon}}(T,I,\vP^T) + \mathcal{L}_{\text{NCS}}(I_{1}, I_{2})
\label{eq:mvsimcon_overall}
\end{dmath}
The design choices in MV-SimCon are methodically made based on the empirical evidence as studied in \cref{sec:ablation}.
\section{Experiments}
\label{sec:experiments}

\subsection{Training and Evaluation Datasets}
\label{sec:datasets}

\paragraph{Training datasets.} The experiments use Google's Conceptual Captions GCC3M \cite{sharma2018conceptual}, GCC12M \cite{changpinyo2021conceptual}, RedCaps12M \cite{desai2021redcaps} and filtered YFCC14M \cite{yfcc100m}. The exact number of samples for these datasets in our version, along with the number of samples in GroupViT \cite{xu2022groupvit} implementation, are reported in \cref{tab:datasets}. Note that the images in these datasets are hosted on a range of sources on the web, where the links change or expire with time. Therefore, the number of samples in our version of the dataset is lower than those in \cite{xu2022groupvit}. To investigate the efficacy of the proposed method, we experiment with different numbers of training samples, starting with $3$ million from GCC3M to $41$ million with all the mentioned datasets combined. Further, we also experiment with different distributions at a similar scale by comparing models trained on GCC12M with ones trained on RedCaps12M.

\begin{table}[H]
\setlength\extrarowheight{1pt}
\begin{center}
\footnotesize
\begin{tabular}{l|l|l|l|l}
    \hline
    \textbf{Dataset} & \textbf{Avg.} & \textbf{\#Samples} &  \textbf{\#Samples } & \textbf{\% diff.}\\
    & \textbf{Length} & \textbf{Ours} & \textbf{\cite{xu2022groupvit}} & \\
    \hline\hline
    GCC3M \cite{sharma2018conceptual} & 10.5 & $2.857$M & $2.891$M & $-1.2\%$\\
    GCC12M \cite{changpinyo2021conceptual} & 22.4 & $10.696$M & $11.156$M & $-4.1\%$\\
    Redcaps12M \cite{desai2021redcaps} & 11.8 & $11.835$M & $11.866$M & $-0.3\%$\\
    YFCC14M \cite{yfcc100m} & 38.4 &$14.611$M & $14.615$M & $-.03\%$\\
    \hline
\end{tabular}
\end{center}
\vspace{-1.5em}
\caption{Datasets used for the training along with the number of samples in our and in GroupViT's \cite{xu2022groupvit} version of the datasets.}
\label{tab:datasets}
\vspace{-1em}
\end{table}

\begin{table*}[t]
\begin{center}
\small
\setlength{\tabcolsep}{6pt}
    \begin{tabular}{ll|lll|l}
    \hline
    \textbf{Loss Function} & 
    \textbf{Training Data} & 
    \textbf{PASCAL VOC} & 
    \textbf{PASCAL Context} & 
    \textbf{COCO} &
    \textbf{Average}\\
    \hline\hline
    
    InfoNCE \cite{xu2022groupvit} & 
    CC3M &
    $16.0$ &
    $7.20$ &
    $6.50$ &
    $9.90$
    \\

    SimCon (Ours) & 
    CC3M &
    $30.4$ \textcolor{blue}{$+14.4\%$} &
    $15.1$ \textcolor{blue}{$+7.90\%$} &
    $12.2$ \textcolor{blue}{$+5.70\%$} &
    $19.2$ \textcolor{blue}{$+9.30\%$}
    \\
    
    MV-SimCon (Ours) &
    CC3M &
    $35.0$ \textcolor{blue}{$+19.0\%$} &
    $17.1$ \textcolor{blue}{$+9.90\%$} &
    $13.4$ \textcolor{blue}{$+6.90\%$} &
    $21.8$ \textcolor{blue}{$+11.9\%$}
    \\
    \hdashline
    
    InfoNCE \cite{xu2022groupvit} & 
    R12M &
    $19.1$ &
    $11.0$ &
    $8.9$ &
    $13.0$\\

    SimCon (Ours) & 
    R12M &
    $37.9$ \textcolor{blue}{$+18.8\%$} &
    $18.1$ \textcolor{blue}{$+7.10\%$} &
    $19.5$ \textcolor{blue}{$+10.6\%$} &
    $25.2$ \textcolor{blue}{$+12.2\%$}
    \\
    
    MV-SimCon (Ours) &
    R12M &
    $40.7$ \textcolor{blue}{$+21.6\%$} &
    $19.1$ \textcolor{blue}{$+8.10\%$} &
    $21.6$ \textcolor{blue}{$+12.7\%$} &
    $27.1$ \textcolor{blue}{$+14.1\%$}
    \\
    \hdashline

    InfoNCE \cite{xu2022groupvit} & 
    CC12M &
    $41.4$ &
    $19.6$ &
    $20.5$ &
    $27.1$
    \\

    SimCon (Ours) & 
    CC12M &
    $47.1$ \textcolor{blue}{$+5.70\%$} &
    $21.3$ \textcolor{blue}{$+1.70\%$} &
    $22.6$ \textcolor{blue}{$+2.10\%$} &
    $30.3$ \textcolor{blue}{$+3.20\%$}
    \\
    
    MV-SimCon (Ours) &
    CC12M &
    $48.9$ \textcolor{blue}{$+7.50\%$}&
    $23.0$ \textcolor{blue}{$+3.40\%$} &
    $23.8$ \textcolor{blue}{$+3.30\%$} &
    $31.9$ \textcolor{blue}{$+4.80\%$}
    \\
    

    




    \hline
\end{tabular}
    \vspace{-1em}
    \caption{Zero-shot semantic segmentation results on PASCAL-VOC \cite{evw+10}, PASCAL Context \cite{mottaghi2014role}, and COCO \cite{lmb+14} measured with mask mIoU (\%) with different training loss functions. Each model is trained independently either on GCC3M \cite{sharma2018conceptual}, RedCaps12M \cite{desai2021redcaps}, or GCC12M \cite{changpinyo2021conceptual} dataset with the same setup. Absolute improvements (\%) over the baseline \cite{xu2022groupvit} are shown in \textcolor{blue}{blue}.
    }
    \label{tab:seg_results_independent_datasets}
    \vspace{-2em}
\end{center}
\end{table*}

\begin{figure*}
\centering
\pgfplotstableread{
epoch	con	mvsimcon	simcon
1	4.49	5.26	4.22
2	5.75	6.75	4.72
3	6.26	12.33	7.94
4	7.11	15.99	9.12
5	9.77	18.76	12.19
6	8.18	21.09	13.7
7	11.45	23.48	16.14
8	10.75	25.09	17.08
9	12.79	26.36	17.93
10	13.31	28.66	17.5
11  15.99	28.53	21.43
12  15.08	28.52	20.9
13  12.90	29.98	23.29
14  12.33	30.65	24.73
15  11.64	30.32	23.98
16  11.96	30.45	25.27
17  12.55	31.15	27.45
18  13.47	32.78	27.11
19  11.38	32.99	27.68
20  12.82	35.02	29.25
21  11.92	33.33	29
22  12.00	33.22	28.56
23  13.53	33.97	29.1
24  13.58	33.54	30.42
25  13.73	33.61	29.91
26  14.69	33.66	30.58
27  14.38	33.46	30.03
28  14.94	34.44	30.49
29  14.90	33.66	30.32
30  15.03	33.76	30.5		
 	}{\yfccLambda}
\begin{tikzpicture}
\begin{axis}[%
	width=0.37\linewidth,
	height=0.25\linewidth,
	xlabel={\small epoch $\rightarrow$},
	ylabel={\small mIoU $\rightarrow$},
	title={GCC3M},
	legend cell align={left},
    legend style={at={(0.17,-0.17)}, cells={anchor=east}, font =\tiny, fill opacity=0.8, row sep=-2.5pt},
   	xtick={1,5,10,15,20,25,30},
   	xmode=linear,
    grid=both,
]
	\addplot[color=red,     solid, mark=*,  mark size=1.0, line width=1.0] table[x=epoch, y expr={\thisrow{con}}] 
	\yfccLambda;
	\addplot[color=black,     solid, mark=square*,  mark size=1.0, line width=1.0] table[x=epoch, y expr={\thisrow{simcon}}] 
	\yfccLambda;
	\addplot[color=blue,     solid, mark=diamond*,  mark size=1.0, line width=1.0] table[x=epoch, y expr={\thisrow{mvsimcon}}] 
	\yfccLambda;

\end{axis}
\end{tikzpicture}
\pgfplotstableread{
epoch	con	mvsimcon	simcon
1	5.98	9.88	5.22
2	7.41	16.15	10.99
3	7.51	21.95	16.2
4	11.06	28.36	15.62
5	12.01	27.75	22.42
6	11.49	31.64	22.25
7	13.61	30.45	25.12
8	14.09	30.87	25.59
9	14.13	33.79	27.87
10	13.61	31.44	31.16
11	15.17	33.05	29.72
12	16.83	35.44	32.12
13	15.44	34.27	30.42
14	15	36.63	34.25
15	16.19	34.88	33.71
16	15.89	35.65	35.48
17	14.57	37.62	34.97
18	16.85	37.65	35.3
19	16.54	37.99	33.95
20	16.43	37.49	36.86
21	16.66	37.92	35.46
22	18.56	38.78	35.64
23	18	38.13	36.69
24	18.29	40.02	38.07
25	18.73	40.19	36.49
26	19.09	38.7	37.27
27	19.08	40.66	37.92
28	18.86	40.25	37.39
29	18.86	39.98	36.71
30	18.86	39.51	37.27 
}{\yfccLambda}
\begin{tikzpicture}
\begin{axis}[%
	width=0.37\linewidth,
	height=0.25\linewidth,
	xlabel={\small epoch $\rightarrow$},
	title={RedCaps12M},
	legend cell align={left},
    legend style={at={(0.17,-0.17)}, cells={anchor=east}, font =\tiny, fill opacity=0.8, row sep=-2.5pt},
   	xtick={1,5,10,15,20,25,30},
   	xmode=linear,
    grid=both,
]
	\addplot[color=red,     solid, mark=*,  mark size=1.0, line width=1.0] table[x=epoch, y expr={\thisrow{con}}]
	\yfccLambda;
	\addplot[color=black,     solid, mark=square*,  mark size=1.0, line width=1.0] table[x=epoch, y expr={\thisrow{simcon}}] 
	\yfccLambda;
	\addplot[color=blue,     solid, mark=diamond*,  mark size=1.0, line width=1.0] table[x=epoch, y expr={\thisrow{mvsimcon}}] 
	\yfccLambda;
\end{axis}
\end{tikzpicture}
\pgfplotstableread{
epoch	con	mvsimcon	simcon
1	6.23	10.72	7.56
2	11.71	23.39	12.79
3	16.89	30.40	20.44
4	22.62	33.66	26.52
5	27.42	36.88	27.84
6	27.8	36.67	30.48
7	28.43	39.66	31.78
8	30.6	40.13	33.31
9	29.93	41.18	35.09
10	32.02	42.07	37.65
11	31.73	43.06	36.51
12	36.07	43.31	37.42
13	35.96	45.04	38.05
14	37.52	45.65	40.09
15	38.09	45.41	40.63
16	37.45	44.97	41.77
17	40.7	46.28	42.34
18	38.54	46.50	42.59
19	38.56	46.25	43.1
20	39.73	48.05	44.19
21	40.22	47.66	44.72
22	37.8	48.18	44.19
23	41.69	48.92	45.05
24	41.32	48.71	45.06
25	40.17	48.16	46.04
26	40.72	48.01	45.58
27	40.33	48.18	47.12
28	41.81	48.66	46.31
29	41.18	48.29	46.31
30	41.41	48.51	46.31	
 	}{\yfccLambda}
\begin{tikzpicture}
\begin{axis}[%
	width=0.37\linewidth,
	height=0.25\linewidth,
	xlabel={\small epoch $\rightarrow$},
	title={GCC12M},
	legend cell align={left},
	legend pos=south east,
    legend style={at={(1,0)}, cells={anchor=east}, font =\small, fill opacity=0.8, row sep=-2.5pt},
   	xtick={1,5,10,15,20,25,30},
   	xmode=linear,
    grid=both,
]
	\addplot[color=red,     solid, mark=*,  mark size=1.0, line width=1.0] table[x=epoch, y expr={\thisrow{con}}]
	\yfccLambda;
	\addlegendentry{InfoNCE};
	\addplot[color=black,     solid, mark=square*,  mark size=1.0, line width=1.0] table[x=epoch, y expr={\thisrow{simcon}}] 
	\yfccLambda;
	\addlegendentry{SimCon};
	\addplot[color=blue,     solid, mark=diamond*,  mark size=1.0, line width=1.0] table[x=epoch, y expr={\thisrow{mvsimcon}}] 
	\yfccLambda;
	\addlegendentry{MVSimCon};
\end{axis}
\end{tikzpicture}
\vspace{-2.5em}
\caption{Zero-shot semantic segmentation results on PASCAL VOC \cite{evw+10} measured with mask mIoU (\%) after each training epoch.}
\vspace{-1em}
\label{fig:epoch_miou}
\end{figure*}
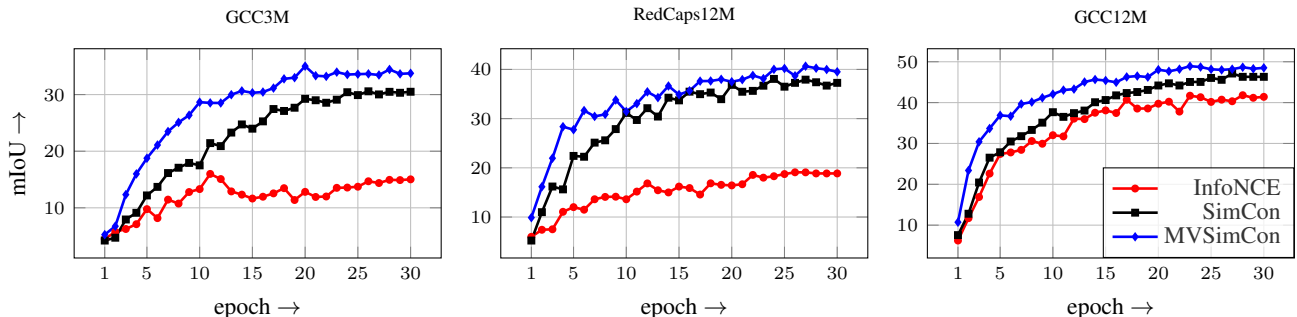

\paragraph{Evaluation datasets.} The proposed approach is evaluated for the task of zero-shot transfer to semantic segmentation on the validation sets of PASCAL VOC \cite{evw+10}, PASCAL Context \cite{mottaghi2014role}, and Microsoft COCO \cite{lmb+14}. They each contain $20$, $59$, and $80$ foreground classes, respectively, with an additional background class. For COCO, following GroupVIT \cite{xu2022groupvit}, the instance segmentation masks from the same class are combined to obtain segmentation masks.

\subsection{Implementation Details}
\label{sec:implementation_details}

\paragraph{Training.} Each model is trained on the specified dataset with a batch size of $2048$ for $30$ epochs with AdamW optimizer \cite{loshchilov2017decoupled}. An initial learning rate of $4e^{-6}$ is linearly warmed up to a maximum learning rate of $1.6e^{-3}$ in the first $2$ epochs. Following warmup, the learning rate is decayed via the cosine schedule \cite{loshchilov2016sgdr}. All the experiments use $8$ NVIDIA-A100 GPUs with $40$GB of memory each. The threshold on intra-modal similarity values for determining positive samples in SimCon and MV-SimCon ($\lambda$ in \cref{eqn:independent_p} and \cref{eqn:joint_p}) is initialized with $0.95$ and is decayed using a step schedule by $0.05$ after the 2nd and the 15th epochs.

\paragraph{Differences with GroupViT.} As noted in \cref{tab:datasets}, our version of these web-based datasets have fewer samples. Furthermore, due to hardware constraints, all the experiments in this paper are conducted with a batch size of $2048$, whereas GroupViT \cite{xu2022groupvit} uses a larger batch size of $4096$. In vision-language pre-training tasks with contrastive learning, the use of larger batch sizes has been shown to give better results \cite{zhai2022lit}. The effect of batch size on the proposed MV-SimCon is studied in \cref{sec:ablation} and the supplementary. With these unavoidable differences in the setup, we compare the proposed approach with our reproduction of GroupViT \cite{xu2022groupvit} results in the exact same setup~\cite{groupvit_github}. 

\paragraph{Discussion on Multi-label loss.} In addition to InfoNCE loss, GroupViT \cite{xu2022groupvit} uses a multi-label contrastive loss function for training. Where the nouns are extracted from the text and fed to a randomly sampled prompt template to construct additional text samples. In the multi-label contrastive loss objective, an image should not only align to the text from the data but also with the auxiliary texts. The use of the multi-label loss function increases the computational requirements as the auxiliary texts are passed through the text encoder. In its default configuration, this generated three auxiliary texts for each sample which restricts the training to datasets that contain long enough texts to extract nouns from, eliminating datasets such as RedCaps12M \cite{desai2021redcaps} due to its short average token length as shown in \cref{tab:datasets}.

In our experiments for the baseline, using multi-label loss increased the computational and GPU memory costs, and either gave similar or worse results. As an example, when trained on GCC12M using InfoNCE loss alone, the model attains mIoU of $41.4\%$ on PASCAL VOC, whereas a model trained with InfoNCE and multi-label loss attains $41.1\%$ (reported in \cite{xu2022groupvit}). Additionally, when trained on $27$ million samples from GCC3M, GCC12M and YFCC14M, the model achieves a mIoU of $50.3\%$ on PASCAL VOC, whereas the performance degrades to $47.5\%$ when the multi-label loss is used. For these reasons, the multi-label loss is not used in our reproduction of the baseline.

\begin{table*}[t]
\begin{center}
\footnotesize
\setlength{\tabcolsep}{2pt}
        \begin{tabular}{c|cc|ccc|lll}
    \hline
    \textbf{Model} &
    \textbf{Arch.} &
    \textbf{Pre-training} & 
    \textbf{Supervision} & 
    \textbf{Zero} & 
    \textbf{TTA} & 
    \textbf{PASCAL} &
    \textbf{PASCAL} &
    \textbf{COCO}\\
    &
    &
    \textbf{Dataset} &
    &
    \textbf{shot} &
    &
    \textbf{VOC} &
    \textbf{Context} &
    \\
    \hline\hline
     
    DeiT \cite{touvron2021training} &
    ViT  &
    ImageNet (1.2M) &
    class &
    \xmark &
    \xmark &
    $53.0$ &
    $35.9$ &
    -
    \\
    \hline
    
    DINO \cite{caron2021emerging} &
    ViT  &
    ImageNet (1.2M) &
    self &
    \xmark &
    \xmark &
    $39.1$ &
    $20.4$ &
    -
    \\
    
    DINO \cite{caron2021emerging} &
    ViT  &
    CC3 + CC12 + Y14 (29M) &
    self &
    \xmark &
    \xmark &
    $37.6$ &
    $22.8$ &
    -
    \\
    
    MoCo \cite{he2020momentum} &
    ViT  &
    ImageNet (1.2M) &
    self &
    \xmark &
    \xmark &
    $34.3$ &
    $21.3$ &
    -
    \\
    
    MoCo \cite{he2020momentum} &
    ViT  &
    CC3 + CC12 + Y14 (29M) &
    self &
    \xmark &
    \xmark &
    $36.1$ &
    $23.0$ &
    -
    \\
    
    \hline
    \color{gray}{InfoNCE} \cite{xu2022groupvit} &
    \color{gray}{GroupViT} &
    \color{gray}{CC3 + CC12 + R12 (27M)} &
    \color{gray}{text} &
    \color{gray}{\ymark} &
    \color{gray}{\xmark} &
    \color{gray}{$50.8$} &
    \color{gray}{$23.7$} &
    \color{gray}{$27.5$}
    \\

    \color{gray}{InfoNCE + Multi-label} \cite{xu2022groupvit} &
    \color{gray}{GroupViT} &
    \color{gray}{CC3 + CC12 + Y14 (29M)} &
    \color{gray}{text} &
    \color{gray}{\ymark} &
    \color{gray}{\xmark} &
    \color{gray}{$52.3$} &
    \color{gray}{$22.4$} &
    \color{gray}{$24.3$}
    \\
    \hdashline
    
    InfoNCE\textsuperscript{\dag} \cite{xu2022groupvit} &
    GroupViT &
    CC3 + CC12 + R12 (27M) &
    text &
    \ymark &
    \xmark &
    $44.7$ &
    $20.0$ &
    $23.4$
    \\

    InfoNCE\textsuperscript{\dag} \cite{xu2022groupvit} &
    GroupViT &
    CC3 + CC12 + Y14 (29M) &
    text &
    \ymark &
    \xmark &
    $50.3$ &
    $21.7$ &
    $24.6$
    \\

    InfoNCE\textsuperscript{\dag} \cite{xu2022groupvit} &
    GroupViT &
    CC3 + CC12 + R12 + Y14 (41M) &
    text &
    \ymark &
    \xmark &
    $50.5$ &
    $20.9$ &
    $23.0$
    \\
    
    \hline

    MV-SimCon\textsuperscript{\dag} &
    GroupViT &
    CC3 + CC12 + R12 (27M) &
    text &
    \ymark &
    \xmark &
    $52.3$ \textcolor{blue}{$+7.60\%$} &
    $\bm{24.5}$ \textcolor{blue}{$+4.50\%$} &
    $27.7$ \textcolor{blue}{$+4.30\%$}
    \\
    

    
    MV-SimCon\textsuperscript{\dag} &
    GroupViT &
    CC3 + CC12 + Y14 (29M) &
    text &
    \ymark &
    \xmark &
    $52.4$ \textcolor{blue}{$+2.10\%$} &
    $22.2$ \textcolor{blue}{$+0.50\%$} &
    $26.6$ \textcolor{blue}{$+2.00\%$}
    \\

    
    MV-SimCon\textsuperscript{\dag} &
    GroupViT &
    CC3 + CC12 + R12 + Y14 (41M) &
    text &
    \ymark &
    \xmark &
    $\bm{53.5}$ \textcolor{blue}{$+3.00\%$} &
    $24.2$ \textcolor{blue}{$+3.30\%$} &
    $\bm{29.9}$ \textcolor{blue}{$+6.90\%$}
    \\




    \hline
    InfoNCE\textsuperscript{\dag} \cite{xu2022groupvit} &
    GroupViT &
    CC3 + CC12 + R12 + Y14 (41M) &
    text &
    \ymark &
    \ymark &
    $53.2$ &
    $22.7$ &
    $24.8$
    \\
    
    MV-SimCon\textsuperscript{\dag} &
    GroupViT &
    CC3 + CC12 + R12 + Y14 (41M) &
    text &
    \ymark &
    \ymark &
    $\bm{58.7}$ \textcolor{blue}{$+5.50\%$} &
    $\bm{26.6}$ \textcolor{blue}{$+3.90\%$} &
    $\bm{33.3}$ \textcolor{blue}{$+8.50\%$}
    \\
    \hline
    \end{tabular}
    \vspace{-1em}
    \caption{Mask mIoU (\%) on PASCAL-VOC \cite{evw+10}, PASCAL Context \cite{mottaghi2014role} and COCO \cite{lmb+14} datasets. Comparisons between zero-shot and fully supervised transfer. Zero-shot "\ymark" indicates transfer to semantic segmentation without any fine-tuning. Absolute improvements (\%) over the baseline \cite{xu2022groupvit} are shown in \textcolor{blue}{blue}. The authors of this paper train all models that are marked with a $\dag$ with the same data and batch size. In \textcolor{gray}{gray} are the models trained by the authors of \cite{xu2022groupvit} with different versions of the data and a higher batch size as noted in \cref{sec:implementation_details}. TTA "\ymark" indicates the use of test time augmentations for inference such as flip, multiple scales, and evaluation at higher resolution.}
    \label{tab:seg_supervised_transfer}
    \vspace{-3em}
\end{center}
\end{table*}

\subsection{Evaluation}
\label{sec:evaluation}

\paragraph{Independent training datasets.}  The results by training the model independently on either GCC3M \cite{sharma2018conceptual}, RedCaps12M \cite{desai2021redcaps} or GCC12M \cite{changpinyo2021conceptual} are shown in \cref{tab:seg_results_independent_datasets}, where improvements are observed with the use of every pre-training dataset and across all evaluation datasets. The proposed SimCon loss function shows a substantial improvement ranging from an average gain of $3.2\%$ to $9.3\%$. The proposed MV-SimCon setup demonstrates additional improvements on top of  SimCon. Note that most of the improvements come from using SimCon with MV-SimCon adding an additional $0.4\%$ to $2.6\%$ to the average performance. While InfoNCE achieves an average performance of $27.1\%$ mIoU when training on GCC12M, it only attains an average performance of $13.0\%$ mIoU when training on RedCaps12M, which is of a similar scale in terms of the number of samples. This result shows that training with the InfoNCE loss is highly sensitive to data distribution. On the other hand, the proposed MV-SimCon attains $31.9\%$ mIoU average performance when trained on GCC12M and $27.1\%$ mIoU average performance when trained on RedCaps12M, which demonstrates the robustness of MV-SimCon across different data distributions at the same scale.

The performance on PASCAL VOC \cite{evw+10} measured after each training epoch is shown in \cref{fig:epoch_miou}. MV-SimCon converges faster than SimCon, while both SimCon and MV-SimCon improve and converge faster than InfoNCE. When training on GCC3M, SimCon and MV-SimCon outperform the final performance of InfoNCE (after $30$ epochs) after only $7$ and $3$ epochs of training, respectively. Similar observations were made while training on RedCaps12M. On GCC12M, MV-SimCon improves the fastest.

\begin{figure*}[t]
\begin{center}
\includegraphics[width=\linewidth]{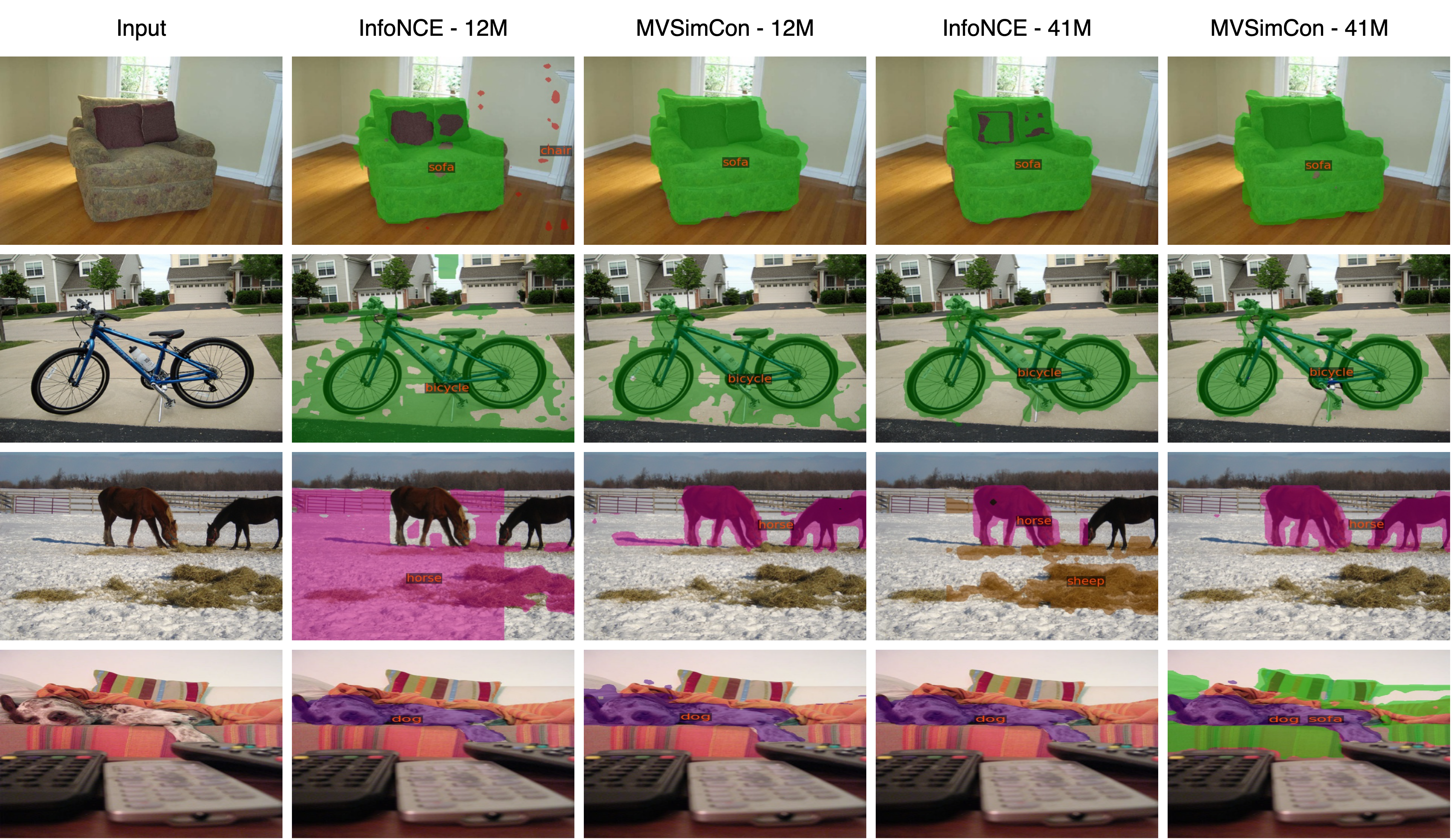}
\vspace{-2em}
\caption{Qualitative results with models trained on $12$ and $41$ million samples. The $12$M setup is trained on GCC12M \cite{changpinyo2021conceptual} dataset, and the $41$M setup is trained on a combination of all datasets in \cref{tab:datasets}. Best viewed in color and by zooming in to see the predicted class.}
\label{fig:qualitative_results}
\vspace{-2em}
\end{center}
\end{figure*}

\paragraph{Combining training datasets.} Results by training the model on different combinations of training datasets, along with comparisons to fully supervised transfer, are shown in \cref{tab:seg_supervised_transfer}. For these experiments, GCC3M and GCC12M are always used and are combined with either RedCaps12M or YFCC14M or both. Under the same training setup, {\em i.e.}, the same batch size and training dataset version, the proposed MV-SimCon consistently outperforms InfoNCE loss on all evaluation datasets. Note that increasing the number of training samples from $29$ million to $41$ million for InfoNCE marginally improves the results on PASCAL VOC and degrades the results on PASCAL Context and COCO. Furthermore, the performance on all evaluation datasets is worse, with $27$ million in training samples. On the other hand, MV-SimCon is more robust to the domain distribution at a comparable number of training samples, and the improvements hold with an increased number of training samples. In comparison to the models trained by the authors of GroupViT \cite{xu2022groupvit} with a higher batch size and more training samples in each dataset, the proposed MV-SimCon demonstrates higher scores with smaller batch size and less data. The last two rows on \cref{tab:seg_supervised_transfer} show results with test time augmentations and higher resolution for inference. Additional results and details are in the supplementary.

\paragraph{Qualitative Results.} Visualization of semantic segmentation predictions from different models is shown in \cref{fig:qualitative_results}. The models trained with $41$ million samples perform better than those trained with $12$ million samples for both loss functions. With the same number of training samples, the grouping with MV-SimCon is better and does not have an overly enlarged mask or holes in the mask. Further, the model trained with InfoNCE misses certain semantic classes completely, an issue which is mitigated to some extent with MV-SimCon. More visualizations, including failure cases, are presented in the supplementary.

\subsection{Effect of design choices in MV-SimCon}
\label{sec:ablation}

\paragraph{Ablation.} The effect of design choices in MV-SimCon are summarized in \cref{tab:ablation}. The experiments were conducted by training on GCC3M and evaluating the models on PASCAL VOC. It can be seen that most of the improvements come from using SimCon loss (row-1 vs. row-2). Naively adding multiple views to SimCon, by aligning the images in each view to the text improves the results marginally (row-2 vs. row-3). Adding the negative cosine similarity (NCS) loss between the two views of the image (\cref{eq:ncs_loss}) improves the results further by $1.2\%$ (row-3 vs. row-4). Adding joint computation of the image positives (\cref{eqn:joint_p}) improves the results by $2.7\%$ (row-4 vs. row-5). Note that joint image positives refer to the setup when the positive samples for images are assigned if the intra-modal similarity is higher than the threshold in either view.

\begin{table}[H]
\setlength\extrarowheight{1pt}
\begin{center}
\footnotesize
\begin{tabular}{c|c|c|c|c}
    \hline
    \textbf{SimCon} & \textbf{Multiple} & \textbf{NCS} &  \textbf{Joint image} & \textbf{PASCAL}\\
    & \textbf{views} & & \textbf{positives} & \textbf{VOC}\\
    \hline\hline
    \xmark & \xmark &  \xmark & \xmark & $16.0$ \\
    \ymark & \xmark &  \xmark & \xmark & $30.4$\\
    \ymark & \ymark & \xmark & \xmark & $31.1$ \\
    \ymark & \ymark & \ymark & \xmark & $32.3$ \\
    \ymark & \ymark & \ymark & \ymark & $35.0$ \\
    \hline
\end{tabular}
\end{center}
\caption{Effect of the design choices in MV-SimCon. All the experiments are conducted by training on GCC3M dataset.}
\label{tab:ablation}
\vspace{-1em}
\end{table}

\begin{figure}
\centering
\input{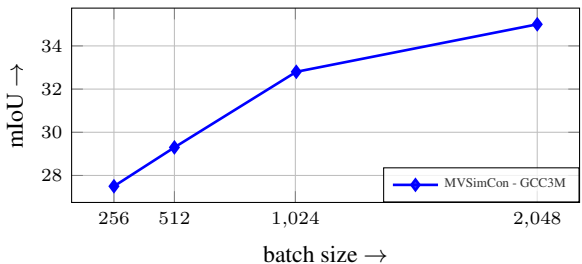}
\begin{tikzpicture}
\begin{axis}[%
	width=\linewidth,
	height=0.5\linewidth,
	xlabel={\small batch size $\rightarrow$},
	ylabel={\small mIoU $\rightarrow$},
	legend cell align={left},
	legend pos=south east,
    legend style={at={(1,0)}, cells={anchor=east}, font =\tiny, fill opacity=0.8, row sep=-4.0pt},
   	xtick={256,512,1024,2048},
   	xmode=linear,
    grid=both,
]
	\addplot[color=blue,     solid, mark=diamond*,  mark size=2.0, line width=1.0] table[x=bs_gcc3, y expr={\thisrow{gcc3m_mvsimcon}}] 
	\yfccLambda;
	\addlegendentry{MVSimCon - GCC3M};
\end{axis}
\end{tikzpicture}
\vspace{-2em}
\caption{Effect of batch size on GCC3M. x-axis is the batch size and y-axis is the mIoU (in \%) on PASCAL VOC dataset.}
\label{fig:batch_size_gcc3m}
\vspace{-2em}
\end{figure}

\paragraph{Effect of batch size.} Results with varying the batch size are shown in \cref{fig:batch_size_gcc3m}. Increasing the batch size improves the results. As noted in \cref{sec:implementation_details}, our main experiments are with a batch size of $2048$, whereas the \cite{xu2022groupvit} uses $4096$. This study justifies the lower results for the baseline \cite{xu2022groupvit} in our experiments than the ones reported by the authors. Based on the trend in \cref{fig:batch_size_gcc3m}, using a higher batch size will improve performance for both our approach and the baseline. However, all the comparisons made in our experiments are fair and were conducted with the same batch size and data.

\section{Conclusions}
\label{sec:conclusions}

A novel loss function termed SimCon is proposed, where an image (text) sample should not only align to the corresponding text (image) but also with the text from samples that are similar in the visual (textual) space. Training is further made robust by combining the SimCon loss with multiple views in the visual domain. The empirical results demonstrate that the use of the proposed MV-SimCon loss function leads to SOTA results on zero-shot semantic segmentation, along with faster convergence.


{\small
\bibliographystyle{ieee_fullname}
\bibliography{egbib}
}

\end{document}